\def\year{2022}\relax
\documentclass[letterpaper]{article} % DO NOT CHANGE THIS

\usepackage{aaai22}  % DO NOT CHANGE THIS
\usepackage{times}  % DO NOT CHANGE THIS
\usepackage{helvet} % DO NOT CHANGE THIS
\usepackage{courier}  % DO NOT CHANGE THIS
\usepackage[hyphens]{url}  % DO NOT CHANGE THIS
\usepackage{graphicx} % DO NOT CHANGE THIS
\usepackage{graphicx}  % DO NOT CHANGE THIS
\usepackage{natbib}  % DO NOT CHANGE THIS AND DO NOT ADD ANY OPTIONS TO IT
\usepackage{caption} % DO NOT CHANGE THIS AND DO NOT ADD ANY OPTIONS TO IT
\DeclareCaptionStyle{ruled}{labelfont=normalfont,labelsep=colon,strut=off} % DO NOT CHANGE THIS
\usepackage{natbib}  % DO NOT CHANGE THIS AND DO NOT ADD ANY OPTIONS TO IT
\usepackage{caption} % DO NOT CHANGE THIS AND DO NOT ADD ANY OPTIONS TO IT

\usepackage{epsfig}
\usepackage{url}

\usepackage{makecell}
\usepackage{mathtools}
\usepackage{comment}

\usepackage{xcolor}

\usepackage{amsmath,amssymb} % define this before the line numbering.
\usepackage{amsmath}
\usepackage[all,cmtip]{xy}

\usepackage[pagebackref,breaklinks]{hyperref}

\usepackage{appendix}
\usepackage{multirow}
\usepackage{bm}

\usepackage{xcolor}
\newcommand{\todo}[1]{[{\bf \color{red} TODO: #1}]}

\newcommand{\yq}[1]{[{\color{green} Yq: #1}]}
\graphicspath{{fig/}}

\usepackage{enumitem}

\newcommand{\Eq}[1]{Eq.~(\ref{eq:#1})}
\newcommand{\eq}[1]{\Eq{#1}}

\newcommand{\fig}[1]{Fig.~\ref{fig:#1}}
\newcommand{\tab}[1]{Table~\ref{tab:#1}}

\usepackage{amsmath}
\DeclareMathOperator{\sign}{sign}
\DeclareMathOperator{\Sign}{Sign}

\newcommand{\model}{\emph{bi-half}\xspace}
\newcommand{\Model}{\emph{Bi-half}\xspace}

\usepackage{graphicx}
\usepackage{amsmath}
\usepackage{amssymb}
\usepackage{booktabs}

\usepackage{xspace}
  \newcommand{\latinphrase}[1]{\textit{#1}}  % always italic
\newcommand{\etal}{\latinphrase{et~al.}\xspace}
\newcommand{\ie}{\latinphrase{i.e.}\xspace}

\usepackage{amsmath,amsfonts,bm}

\def\eqref#1{equation~\ref{#1}}
\def\1{\bm{1}}

\DeclareMathAlphabet{\mathsfit}{\encodingdefault}{\sfdefault}{m}{sl}
\SetMathAlphabet{\mathsfit}{bold}{\encodingdefault}{\sfdefault}{bx}{n}

\makeatletter
\newif\if@restonecol
\makeatother

\usepackage[linesnumbered,ruled,vlined]{algorithm2e}%[ruled,vlined]{
\usepackage{algpseudocode}
\usepackage{amsmath}
\title{Equal Bits: Enforcing Equally Distributed Binary Network Weights}
\author{
    Yunqiang Li\equalcontrib, Silvia L. Pintea\equalcontrib \ and
Jan C. van Gemert
}
\affiliations{
   Computer Vision Lab, Delft University of Technology, \\
   Delft, Netherlands\\
}

\usepackage{bibentry}
\begin{document}

%% omit the following lines up until the closing ``}''.
%% Additional authors and addresses can be added with ``\and'',
%% just like the second author.
%% To save space, use either the email address or home page, not both
%\and
%\maketitle

% Authors must not appear in the submitted version. They should be hidden
% as long as the \iclrfinalcopy macro remains commented out below.
% Non-anonymous submissions will be rejected without review.

% The \author macro works with any number of authors. There are two commands
% used to separate the names and addresses of multiple authors: \And and \AND.
%
% Using \And between authors leaves it to \LaTeX{} to determine where to break
% the lines. Using \AND forces a linebreak at that point. So, if \LaTeX{}
% puts 3 of 4 authors names on the first line, and the last on the second
% line, try using \AND instead of \And before the third author name.

% \newcommand{\fix}{\marginpar{FIX}}
% \newcommand{\new}{\marginpar{NEW}}
%\iclrfinalcopy % Uncomment for camera-ready version, but NOT for submission.

\maketitle

%%%%%%%%%%%%%%%%%%%%%%%%%%%%%%%%%%%%%%%%%%%%%%%%%%%%%%%%%%%%%%%%%%%%%%%%%%%%%%
%%%% Submissions may consist of up to 7 pages of technical content plus up to 2 additional pages solely for references
\begin{abstract}
% Binary is efficient
Binary networks are extremely efficient as they use only two symbols to define the network: $\{+1,-1\}$.
One can make the prior distribution of these symbols a design choice. 
% Equal priors good for maximum entropy
The recent IR-Net of Qin \etal argues that imposing a Bernoulli distribution with equal priors (equal bit ratios) over the binary weights leads to maximum entropy and thus minimizes information loss.
However, %while still using the $\sign$ function to binarize the weights,
prior work cannot precisely control the binary weight distribution during training, and therefore cannot guarantee maximum entropy.
% We propose optimal transport
Here, we show that quantizing using optimal transport can guarantee any bit ratio, including equal ratios.
We investigate experimentally that equal bit ratios are indeed preferable and show that our method leads to optimization benefits.
We show that our quantization method is effective when compared to state-of-the-art binarization methods, even when using binary weight pruning. 
Our code is available at 
\href{https://github.com/liyunqianggyn/Equal-Bits-BNN}{{\textcolor[rgb]{1.00,0.00,1.00}{\texttt{https://github.com/liyun
qianggyn/Equal-Bits-BNN}}}}.

\end{abstract}

%Binary networks are extremely efficient as multiplications and additions are replaced with bit shifts.   
%Yet, binarizing real-valued networks reduces representational power: a real-valued weight is reduced to a binary weights with the only two choices of `${-}1$' or `${+}1$'. 
%We make the observation that pruning binary weights adds the value `$0$' as an additional symbol and thus offers more possible solutions.
%We use a fixed pruning rate that allows us to explicitly control the sparsity of the network, which situates our model in-between the binary and ternary models in terms of bit-width.
%The fixed pruning rate may lead to an unbalanced network in terms of binary wright distribution.  
%We ensure a uniform prior over the binary weights, which is desirable from an information theory perspective, by explicitly modeling the weights as a Bernoulli distribution with equal priors. This leads to a regularization of the problem space where these solutions are preferred. We jointly map the real-valued distribution to the Bernoulli distribution and prune the binary weights, thus finding a subnetwork that performs better than the original network. On 3 datasets and 11 architectures we show compact models with good accuracy comparing favorably to the state-of-the-art. 

\section{Introduction}

% Binary nets are great
%We focus on deep binary networks to replace expensive operations such as multiplications and additions with fast bit operations.
Binary networks allow compact storage and swift computations by limiting the network weights to only two bit symbols $\{-1, +1\}$.
In this paper we investigate weights priors before seeing any data: is there a reason to prefer predominantly positive bit weights? Or more negative ones? Or is equality preferable? Successful recent work \cite{qin2020forward} argues that a good prior choice is to have an equal bit-ratio: i.e. an equal number of $+1$ and $-1$ symbols in the network.
This is done by imposing an equal prior under the standard Bernoulli distribution \cite{qin2020forward,peters2018probabilistic,zhou2016dorefa}.
Equal bit distributions minimize  information loss and thus maximizes entropy, showing benefits across architectures and datasets~\cite{qin2020forward}. 
However, current work cannot add a hard constraint of making symbol priors exactly equal, and  therefore cannot guarantee maximum entropy.

% We propose to explicitly control the distribution and this has 2 benefits: (1) we can test any bit ration, (2) we can reduce the solution space
Here, we propose a method to add a hard constraint to binary weight distribution, offering precise control for any desired bit ratio, including equal prior ratios. 
We add hard constraints in the standard quantization setting \cite{bulat2019xnornet,qin2020forward,rastegari2016xnor} making use of real-valued latent weights that approximate the binary weights. 
We quantize these real-valued weights by aligning them to any desired prior Bernoulli distribution, which incorporates our preferred binary weight prior.
Our quantization uses optimal transport~\cite{villani2003topics} and can guarantee any bit ratio.
Our method makes it possible to experimentally test the hypothesis in \cite{qin2020forward} that equal bit ratios are indeed preferable to other bit ratios.
We baptize our approach with equal bit ratios: \model. 
Furthermore, we show that enforcing equal priors using our approach leads to optimization benefits by reducing the problem search-space and avoiding local minima.

% Contributions
We make the following contributions:
(i) a binary network optimization method based on optimal transport;
(ii) exact control over weight bit ratios;
(iii) validation of the assumption that equal bit ratios are preferable;
(iv) optimization benefits such as search-space reduction and good minima;
(v) favorable results compared to the state-of-the-art, and can ensure half-half weight distribution even when pruning is used. 
\begin{table*}
	\centering
	\renewcommand{\arraystretch}{1.2}
	\resizebox{1\linewidth}{!}{
	 \setlength{\tabcolsep}{0.6mm}{
	   	\begin{tabular}{l|l|l|l|l}
    	\toprule
        & \textbf{\textcircled{a} $\Sign$, no filter statistics} &
        \textbf{\textcircled{b} $\Sign$, filter statistics}  &
        \textbf{\textcircled{c} Flip, filter statistics} &
        \textbf{\textcircled{d} Flip, no filter statistics}\\
        & \cite{rastegari2016xnor}
        & \cite{qin2020forward}
        & Our \model
        & \cite{helwegen2019latent}\\ \midrule	

		Initialization	& Gradient $g$;
		                & Gradient $g$;
                        & Gradient $g$; Latent weight $w$;
                        & Gradient $g$; \\
                        & Latent weight $w$;
                        & Latent weight $w$;
                        & Threshold dependent on $w$;
                        & Predefined threshold $\tau$; \\
 		\midrule	
	    Binarization, $b$ 	& $b\leftarrow \text{sign}(w)$
	                        & $b\leftarrow \text{sign}\left(\frac{w - \text{avg}(w)}{\text{std}(w - \text{avg}(w))}\right)$
	                        & $b\leftarrow \text{flip}(b), \  \text{if} \left\{
	                        \begin{array}{ll}
                                \text{ rank}(w)< \frac{D}{2},\text{ and} \text{ rank}(w - \alpha g) \ge \frac{D}{2} \\
                                \text{ rank}(w) \ge \frac{D}{2},\text{ and} \text{ rank}(w - \alpha g) < \frac{D}{2} \\
                            \end{array} \right.$
		  	                & $b\leftarrow \text{flip}(b), \text{if} \left\{
	                            \begin{array}{ll}
                                    \tau < |g|,\text{ and} \\
                                    \text{sign}(g) =  \text{sign}(b) \\
                                \end{array} \right.$ \\
    \bottomrule
	\end{tabular}
	}}
	\caption{\small
    \textbf{Optimization perspectives.}
    (a) Classical binarization methods tie each binary weight $b$ to an associated real-valued latent variable $w$, and quantize each weight by only considering its associated real-valued by using the $\sign$ function.
    (b) Rather than updating the weights independent of each other, recent work uses filter-weight statistics when updating the binary weights.
    (c) Our proposed optimization method does not focus on using the $\sign$ function, but rather flips the binary weights based on the distribution of the real weights, thus the binary weight updates depend on the statistics of the other weights through the rank of $w$.
    (d) Recent work moves away from using the $\sign$ of the latent variables, and instead trains the binary network with bit sign flips, however they still consider independent weight updates.
    }
	\label{tab:related}
\end{table*}

%-----------------------------------------------------------------------------------
\section{Related Work}
For a comprehensive survey on binary networks, see~\cite{qin2020binary}.
In \tab{related} we show the relation between our proposed method and pioneering methods, that are representatives of their peers, in terms of the binarization choices made.
The XNOR method (\tab{related}(a)) was the first to propose binarizing latent real-valued weights using the $\sign$ function~\cite{rastegari2016xnor}.
Rather than making each binary weight depend only on its associated real-value weight or gradient value, IR-Net~\cite{qin2020forward} (\tab{related}(b)) is a prototype method that uses filter-weight statistics to update each individual binary weight.
Here, we also use filter-weight statistics to update the binary weights, however similar to~\cite{helwegen2019latent} \tab{related}(d)) we do not rely on the $\sign$ function for binarization, but instead use binary weight flips. This is a natural choice, as flipping the sign of a binary weight is the only operation one can apply to binary weights.

%-----------------------------------------------------------------------------------
\medskip\noindent \textbf{Sign versus bit flips.}
The front-runners of binary networks are BinaryConnect~\cite{courbariaux2015binaryconnect} and XNOR~\cite{rastegari2016xnor} and rely on auxiliary real weights and the $\sign$
function to define binary weights.
These works are subsequently extended with focus on the scaling factors in XNOR++~\cite{bulat2019xnornet} and BNN+~\cite{darabi2018bnn+}, while HWGQ~\cite{li2017performance} uses the $\sign$ function recursively for binarization.
Bi-Real~\cite{liu2018bi} also uses the $\sign$ function for binarization and analyzes better approximations of the gradient of the $\sign$ function.
From a different perspective, recent work tries to sidestep having to approximate the gradient of the $\sign$ function, and uses bit flips to train binary networks~\cite{helwegen2019latent}.
The sign of the binary weights can be flipped based on searchable~\cite{yang2020searching} or learnable thresholds \cite{liu2020reactnet}.
Here, we also rely on bit flips based on a dynamic thresholding of the real weights, entailed by our optimal transport optimization strategy.

\medskip\noindent \textbf{Using filter statistics or not.}
Commonly, binarization methods define each binary weight update by considering only its associated value in the real-valued latent weights \cite{bulat2019xnornet,li2017performance,rastegari2016xnor,liu2018bi} or in the gradient vector \cite{helwegen2019latent}.
However, binary weights can also be updated using explicit statistics of the other weights in the filter \cite{lin2017towards} or implicitly learned through a function \cite{han2020training}.
The real-valued filter statistics are used in IR-Net \cite{qin2020forward} to enforce a Bernoulli distribution with equal priors.
In hashing, \cite{li2020deep} designed a parameter-free coding layer to maximize hash channel capacity by shifting the network output with median.
Similarly, our optimal transport optimization leads to ranking the real weights, and therefore making use of the statistics of the real-weights in each filter.

%-----------------------------------------------------------------------------------
\medskip\noindent \textbf{Network pruning.}
Pruning has been shown to improve the
efficiency of deep networks~\cite{frankle2020pruning,huang2018learning,lin2017runtime,xiao2019autoprune,ye2020good}.
However, the reason why pruning can bring improvements remains unclear in real-valued networks.
It is commonly believed~\cite{evci2019rigging,frankle2018lottery,malach2020proving,zhou2019deconstructing} that finding the ``important" weight values is crucial for retraining  a small pruned model. Specifically, the ``important" weight values are  inherited~\cite{han2016deep} or re-winded~\cite{frankle2018lottery} from a large trained model.
In contrast, \cite{liu2018rethinking} claims that the selected important weights are typically not useful for the small pruned model, while the pruned architecture itself is more relevant.
The lottery-ticket idea has recently been applied to binary networks \cite{diffenderfer2021multi}.
Here, we show that having equal $+1$ and $-1$ ratios is also optimal when the networks rely on pruning and that our optimal transport optimization can easily be adapted to work with methods using pruning.

\section{Binarizing with optimal transport}
\label{methods}
\subsection{Binary weights}
We define a binary network where the weights $\mathbf{B}$ take binary values $\{1,-1\}^D$. 
The binary weights $\mathbf{B}$ follow a Bernoulli distribution $\mathbf{B} \sim \text{Be}(p_{pos})$, describing the probabilities of individual binary values $b \in \{-1,1\}$ in terms of the hyperparameters $p_{pos}$ and $p_{neg}$:
\begin{equation}
    p(b)= \text{Be}(b \mid p_{pos}) = \left\{
        \begin{array}{ll}
              p_{pos} & \text{ if } b=+1 \\
              p_{neg} = 1-p_{pos}, & \text{ if } b=-1\\
        \end{array} 
    \right. 
    \label{eq:bern}
\end{equation}
To be consistent with previous work, we follow XNOR-Net~\cite{rastegari2016xnor} and apply the binary optimization per individual filter.

Because the matrix $\mathbf{B}$ is discrete, we follow~\cite{courbariaux2015binaryconnect,rastegari2016xnor} by using real-valued latent weights $\mathbf{W}$ to aid the training of discrete values, where each binary weight in $\mathbf{B}$ has an associated real-valued weight in $\mathbf{W}$.
In the forward pass we quantize the real-valued weights $\mathbf{W}$ to estimate the matrix $\mathbf{B}$.
Then, we use the estimated matrix $\mathbf{B}$ to compute the loss, and in the backward pass we update the associated real-valued weights $\mathbf{W}$.

%------------------------------------------------------------
\subsection{Optimal transport optimization}
The optimization aligns the real-valued weight distribution $\mathbf{W}$ with the prior Bernoulli distribution in \eq{bern} and quantizes the real-valued weights $\mathbf{W}$ to $\mathbf{B}$.

The empirical distribution  $\mathbb{P}_w$ of the real-valued variable $\mathbf{W} \in \mathbb{R}^D$ and the empirical distribution $\mathbb{P}_b$ for the discrete variable $\mathbf{B}$ can be written as:
\begin{equation}
    \mathbb{P}_w = \sum_{i=1}^D p_i \delta_{w_i}, \ \ \  \mathbb{P}_b = \sum_{j=1}^{2} q_j \delta_{b_j},
    \label{rewasserp}
\end{equation}
where $\delta_\mathbf{x}$ is the Dirac function at location $\mathbf{x}$.
The $p_i$ and $q_j$ are the probability mass associated to the corresponding distribution locations $w_i$ and $b_j$, where $\mathbb{P}_b$ has only 2 possible locations in the distribution space $\{-1,1\}$.

To align $\mathbb{P}_w$ with the Bernoulli  prior $\mathbb{P}_b$ in \eq{bern} 
we use optimal transport (OT)~\cite{villani2003topics} which minimizes the cost of moving the starting distribution $\mathbb{P}_w$  to the target distribution $\mathbb{P}_b$. Because
$\mathbb{P}_w$ and $\mathbb{P}_b$ are only accessible through a finite set of values, the corresponding optimal transport cost is:
\begin{equation}
    \pi_0= \mathop{\min}_{\pi \in \Pi(\mathbb{P}_w, \mathbb{P}_b)} 	\left \langle \pi, \mathcal{C}\right \rangle_F,
    \label{rewasserp0}
\end{equation}
where $\Pi(\mathbb{P}_w, \mathbb{P}_b)$ is the space of the joint probability with marginals $\mathbb{P}_w$ and $\mathbb{P}_b$, and $\pi$ is the general probabilistic coupling that indicates how much mass is transported to push distribution $\mathbb{P}_w$  towards the distribution $\mathbb{P}_b$.
The $\left \langle ., .\right \rangle_F$ denotes the Frobenius dot product,  and $\mathcal{C} \geq 0$ is the cost function matrix whose element  $\mathcal{C}(w_i, b_j)$ denotes the cost of moving a probability mass from location $w_i$ to location $b_j$ in distribution space.
When the cost is defined as a distance, the OT becomes the Wasserstein distance.
We minimize the 1-Wasserstein distance between $\mathbb{P}_w$ and $\mathbb{P}_b$.
This minimization has an elegant closed-form solution based on simply sorting.
For a continuous-valued weights vector $\mathbf{W} \in \mathbb{R}^D$, we first sort the elements of $\mathbf{W}$, and then assign the top $p_{pos} D$ elements to $+1$, and the bottom $(1-p_{pos}) D$ portion of the elements to $-1$:
\begin{equation}
    \mathbf{B}= \mathrm{\pi_0}\ (\mathbf{W}) =
    \begin{cases}
    +1,  & \mathrm{ top \ p_{pos} D \ of \ sorted}\ \mathbf{W}\\
    -1, & \mathrm{ bottom \ (1-p_{pos}) D \ of \ sorted}\ \mathbf{W}\\
    \end{cases}
    \label{eq:behalf1}
\end{equation}
Rather than using the $\sign$ function to define the binarization, we flip the binary weights based on the distribution of $\mathbf{W}$.
Thus the flipping of a binary weight depends on the distribution of the other binary weights through $\mathbf{W}$, which is optimized to be as close as possible to $\mathbf{B}$.

When applying our method in combination with pruning as in \cite{diffenderfer2021multi}, we first mask the binary weights  $\mathbf{B}^\prime = \mathbf{M}\odot\mathbf{B}$ with a mask $\mathbf{M} \in \{0,1\}^D$. This leads to a certain percentage of the weights being pruned.
Subsequently, we apply the \eq{behalf1} to the remaining non-pruned weights, where $D$ in \eq{behalf1} become the $L_1$ norm of the mask, $|\mathbf{M}|$. 

%--------------------------------------------------------------------------------------
\subsection{\Model: Explicitly controlling the bit ratio} 
Our optimal transport optimization allows us to enforce a hard constraint on precise bit ratios by varying the $p_{pos}$ value.
Therefore, we can test a range of prior binary weight distributions. 

Following \cite{qin2020forward}, a good prior over the binary weights is one maximizing the entropy. 
Using optimal transport, we maximize the entropy of the binary weights by setting the bit ratio to half in \eq{behalf1}:
\begin{equation}
        p^*_{pos} = \text{argmax}_{p_{pos}} H(\mathbf{B} \sim \text{Be}(p_{pos})) = \frac{1}{2},
    \label{eq:bihalf2}
\end{equation}
where $H(\cdot)$ denotes the entropy of the binary weights $\mathbf{B}$.
We dub this approach \model.
Unlike previous work \cite{qin2020forward}, we can guarantee equal symbol distributions and therefore maximum entropy throughout the complete training procedure.

%--------------------------------------------------------------------------------------
\smallskip\noindent \textbf{Initialization and scaling factor.}
We initialize the real-valued weights using Kaiming normal~\cite{he2015delving}. 
The  binary weights are initialized to be equally distributed per filter according to \eq{bihalf2}. 
To circumvent exploding gradients, we use one scaling factor $\alpha$ per layer for the binary weights to keep the activation variance in the forward pass close to~$1$. 
Based on the ReLU variance analysis in~\cite{he2015delving} it holds that $\frac{1}{2} D\cdot Var(\alpha{\mathbf{B}}) = 1$, where $D$ is the number of connections and $\mathbf{B}$ are our binary weights.
$\mathbf{B}$ is regularized to a \model distribution, thus $Var(\mathbf{B}) = 1$, which gives $\alpha = \sqrt{2/D}$.

To better clarify, for an $L$-layer network with input data $y_1$ standardized to $Var[y_1] = 1$, where the variance of each binary layer $l$ is $Var[{\mathbf{B}_l}]=1$, and $D_l$ is the number of connections in that layer: 
\emph{i)} Without the scaling, the output variance is $ Var[y_L] = Var[y_1]\prod_{l = 2}^{L} \frac{D_l}{2}Var[{\mathbf{B}_l}] = \prod_{l = 2}^{L} \frac{D_l}{2}$. 
Typically $D_l$ is large, leading to exploding gradients;
\emph{ii)} With the scaling, we scale ${\mathbf{B}_l}$ by $\alpha = \sqrt{2/D_l}$, leading to $ Var[y_L] = Var[y_1]\prod_{l = 2}^{L} \frac{D_l}{2}  Var(\alpha{\mathbf{B}_l}) = 1$ which stabilizes learning.

\begin{figure*}[t]
    \begin{tabular}{c@{\hskip 0.3in}c}
    \includegraphics[width=0.32\textwidth]{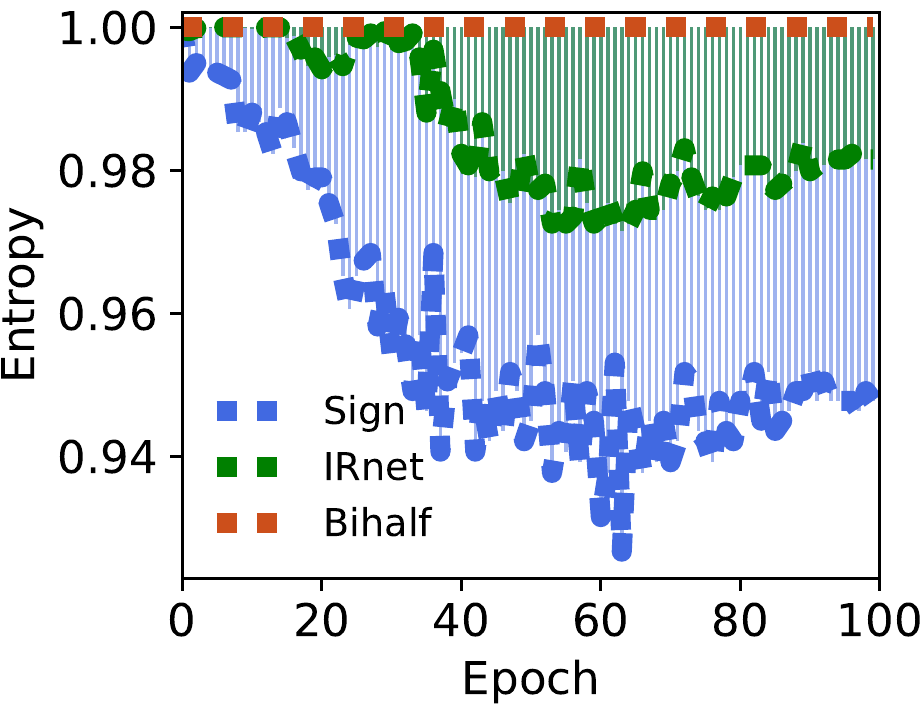} &
    \includegraphics[width=0.6\textwidth]{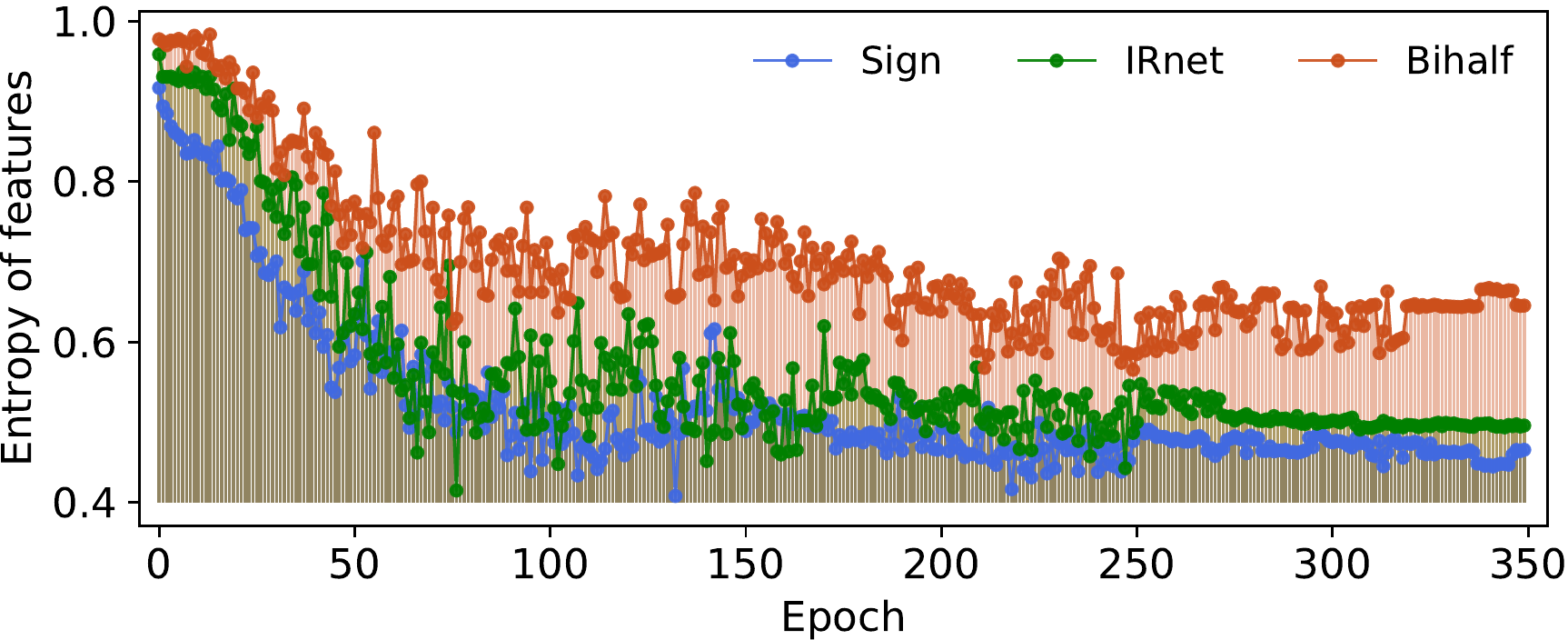} \\
    (a) Binary weight entropy on Conv2. &
    (b) Activation entropy on ResNet-18. \\
    \end{tabular}
    \caption{\small
    \textbf{Hypothesis: bi-half maximizes the entropy.}
    Entropy of binary weights and activations.
    We compare our bi-half method to $sign$ \cite{rastegari2016xnor} and IR-Net \cite{qin2020forward}.
    (a) Entropy of the binary weights during training for Conv2 on Cifar10.
    (b) Entropy of the network activations for ResNet-18 on Cifar100.
    Our \model model can guarantee maximum entropy during training for the
    binary weight distribution and it is able to better maximize the entropy of the activations.
    }
    \label{fig:bit_epochs}
\end{figure*}

%---------------------------------------------------------------------------------------
\section{Experiments}
\noindent \textbf{Datasets and implementation details.}
We evaluate on Cifar-10, Cifar-100~\cite{krizhevsky2009learning} and ImageNet~\cite{deng2009imagenet}, for a number of network architectures.
Following~\cite{frankle2018lottery,ramanujan2020s} we evaluate 4 shallow CNNs: Conv2, Conv4, Conv6, and Conv8 with 2/4/6/8 convolutional layers.
We train the shallow models on Cifar-10 for 100 epochs, with weight decay $1e^{-4}$, momentum 0.9, batch size 128, and initial learning rate 0.1 using a cosine learning rate decay~\cite{loshchilov2016sgdr}.
Following~\cite{qin2020forward} we also evaluate their ResNet-20 architecture and settings on Cifar-10.
On Cifar-100, we evaluate our method on 5 different models including VGG16~\cite{simonyan2014very},  ResNet18~\cite{he2016deep}, ResNet34~\cite{he2016deep}, InceptionV3~\cite{szegedy2016rethinking}, ShuffleNet~\cite{zhang2018shufflenet}.
We train the Cifar-100 models for 350 epochs using SGD with weight decay $5e^{-4}$, momentum 0.9, batch size 128, and initial learning rate 0.1 divided by 10 at epochs 150, 250 and 320.
For ImageNet we use ResNet-18 and ResNet-34 trained for 100 epochs using SGD with momentum $0.9$, weight decay $1e^{-4}$, and batch size 256.
Following ~\cite{liu2018bi,qin2020forward}, the initial learning rate is set as 0.1 and we divide it by 10 at epochs 30, 60, 90.
All our models are trained from scratch without any pre-training.
For the shallow networks we apply our method on all layers, while for the rest we follow~\cite{liu2018bi,qin2020forward}, and apply it on all convolutional and fully-connected layers except the first, last and the downsampling layers.
% We share all code\footnote{Link to our github source code omitted for review}.

%---------------------------------------------------------------------------------------
\begin{figure}[tp!]
    \centering
    \begin{tabular}{c@{\hskip 1in}c}
    \includegraphics[width=0.46\textwidth]{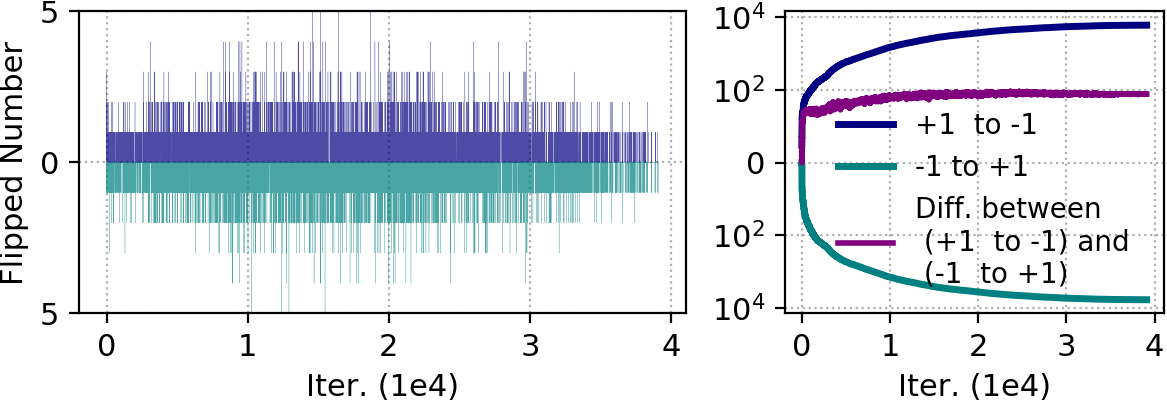} & \\
    \small {(a) $\Sign$ has uneven flips by independently updating binary weight.} & \\
    \includegraphics[width=0.46\textwidth]{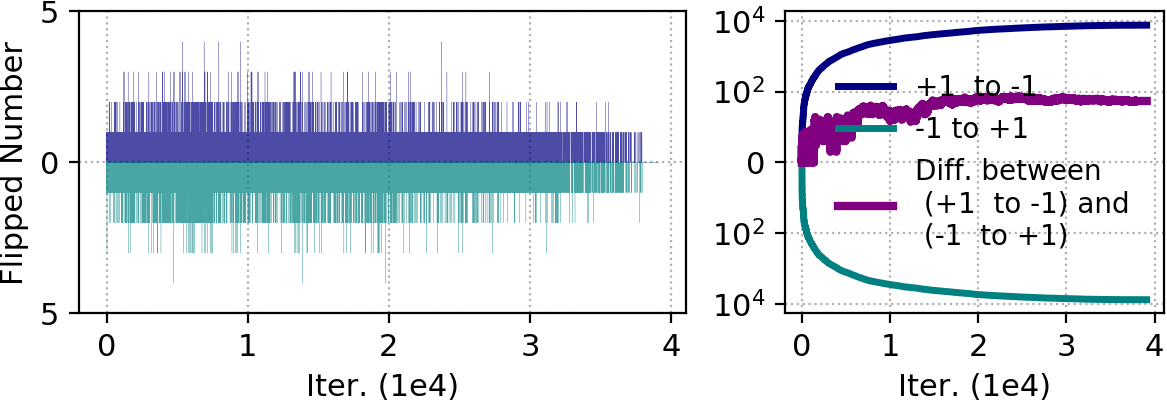} & \\
    \small {(b) IR-Net has uneven flips by balancing the latent weights.} &  \\
    \includegraphics[width=0.46\textwidth]{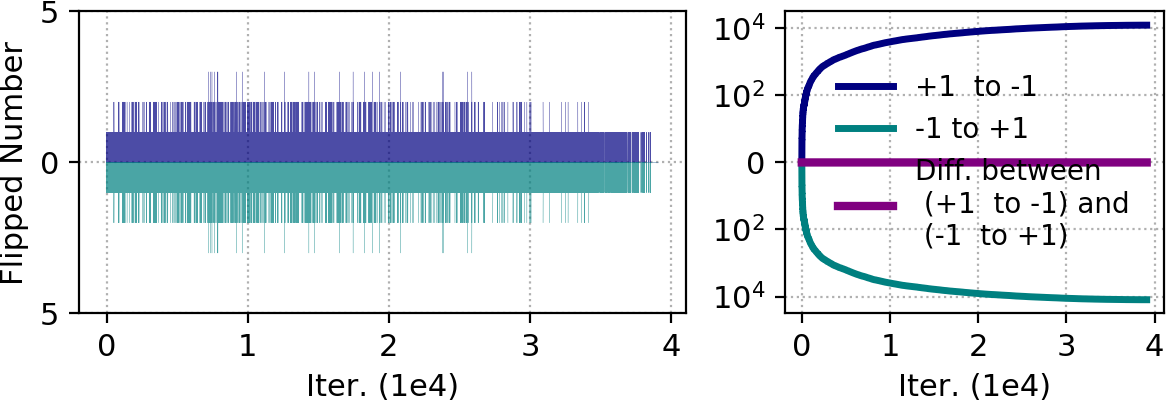} & \\
    \small {(c) \Model has even flips by precisely controlling the flipping. } & \\
    \end{tabular}
    \caption{\small \textbf{Hypothesis: \model maximizes the entropy.}
    Bit flips during training.
    We compare the bit flips during training in our \model with the $\sign$~\cite{rastegari2016xnor} and IR-Net~\cite{qin2020forward} on the Conv2 network on Cifar-10. The x-axis shows the training iterations.
    \emph{Left:} Bit flips during training to +1 (dark blue) or to -1 (cyan).
    \emph{Right:} Accumulated bit flips over the training iterations, as well as the difference between the bit flips from ($+1$ to $-1$) and the ones from ($-1$ to $+1$).
    In contrast to $\sign$ and IR-Net, our \model method can guarantee an equal bit ratio.
    }
    \label{fig:evenflip}
\end{figure}

%------------------------------------------------------------------------------------------
\subsection{\textbf{Hypothesis}: \Model maximizes the entropy}
Here we test whether our proposed \model model can indeed guarantee maximum entropy and therefore an exactly equal ratio of the $-1$ and $+1$ symbols.
\fig{evenflip} shows the bit flips performed in our proposed \model method during training when compared to two baselines: $\sign$~\cite{rastegari2016xnor} and IR-Net \cite{qin2020forward}.
We train a Conv2 network on Cifar-10 and plot the flips of binary weights in a single binary weight filter during training.
The binary weights are initialized to be equal distributed (half of the weights positive and the other half negative).
The classical $\sign$ method~\cite{rastegari2016xnor} in \fig{evenflip}(c) binarizes each weight independent of the other weights, therefore during training the flips for ($+1$ to $-1$) and ($-1$ to $+1$) are uneven.
The recent IR-Net~\cite{qin2020forward} in \fig{evenflip}(b) balances the latent weights by using their statistics to obtain evenly distributed binary weight values.
However, it can not guarantee evenly distributed binary weights throughout training.
Our \model model in \fig{evenflip}(c) updates the binary weight based on the statistics of the other weights.
For our method the binary weights are evenly flipped during training, offering exact control of bit weight ratios.

\fig{bit_epochs}(a) shows the binary weights entropy changes during training on Conv2 when compared to $\sign$~\cite{rastegari2016xnor} and IR-Net~\cite{qin2020forward}.
IR-Net aims to maximize entropy by subtracting the mean value of the weights, yet, this is not exact. In contrast, we maximize the information entropy by precisely controlling the binary weight distribution.
In \fig{bit_epochs}(b) we show the entropy of the binary activations.  Adjusting the distribution of binary weights retains the information in the binary activation. For our \model method, the binary activation of each channel is close to the maximum information entropy under the Bernoulli distribution.
% \fig{bit_epochs}(c) visualizes the distribution of latent real-valued weights before binarization.
% The latent weight distribution is not symmetric, and therefore simply subtracting the mean value as in IR-Net cannot achieve an equal prior distribution \jvg{I do not see that IR-Net is asymmetric}.

%---------------------------------------------------------------------------------------

\begin{figure*}[t]
    \centering
    \begin{tabular}{c@{}c@{}c@{}c@{}c@{}}
    \includegraphics[width=0.198\textwidth]{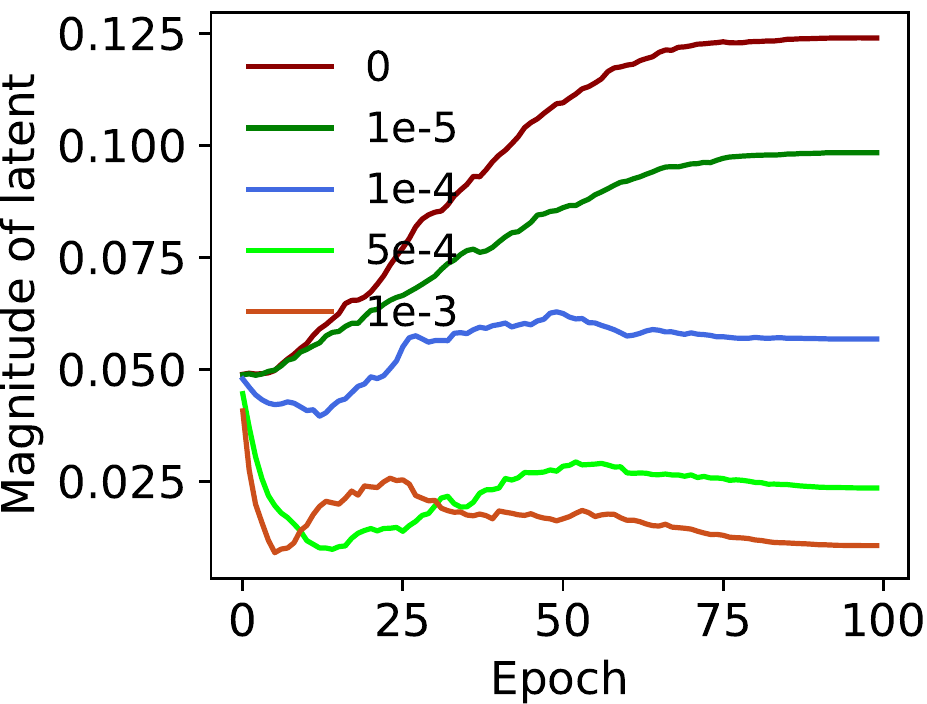} &  \
    \includegraphics[width=0.192\textwidth]{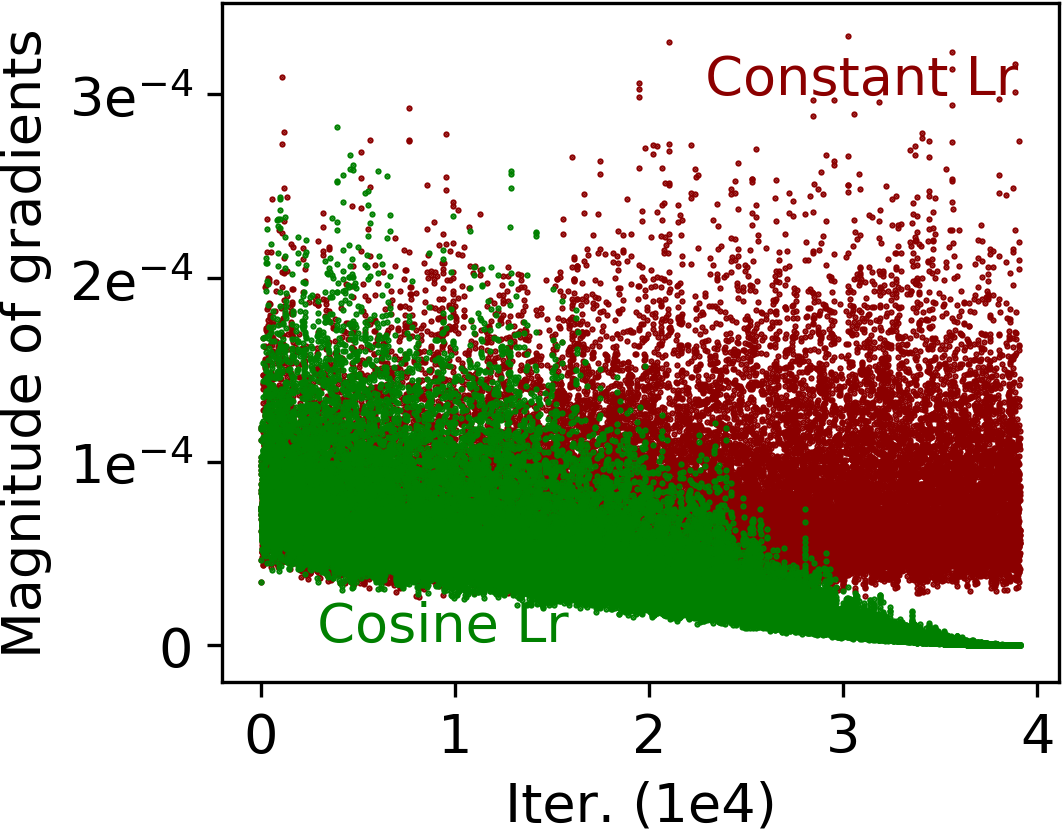}   & \
    \includegraphics[width=0.196\textwidth]{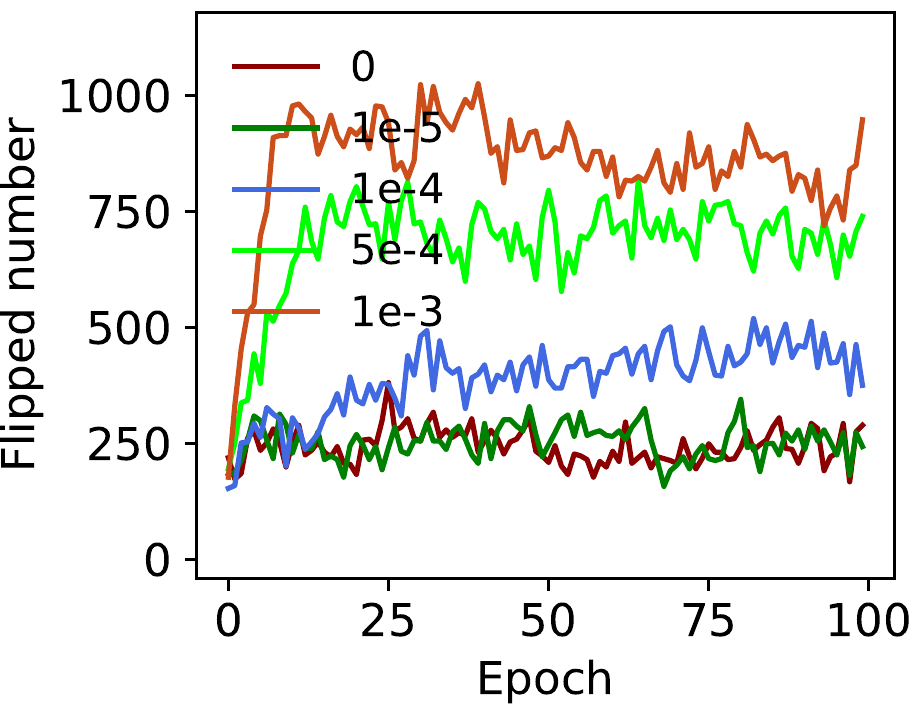} &  \
    \includegraphics[width=0.196\textwidth]{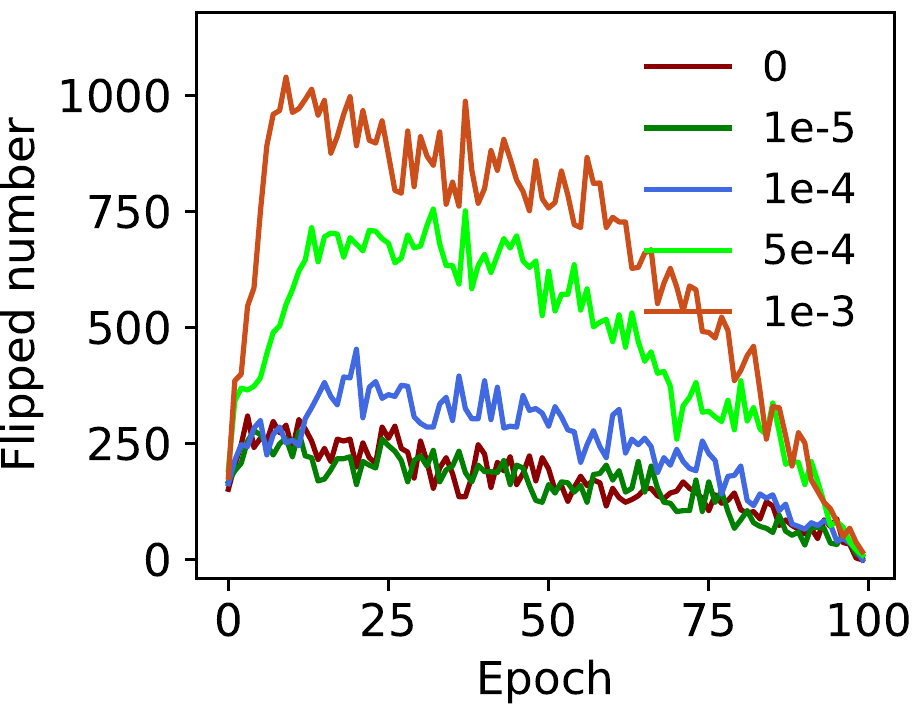} &  \
    \includegraphics[width=0.194\textwidth]{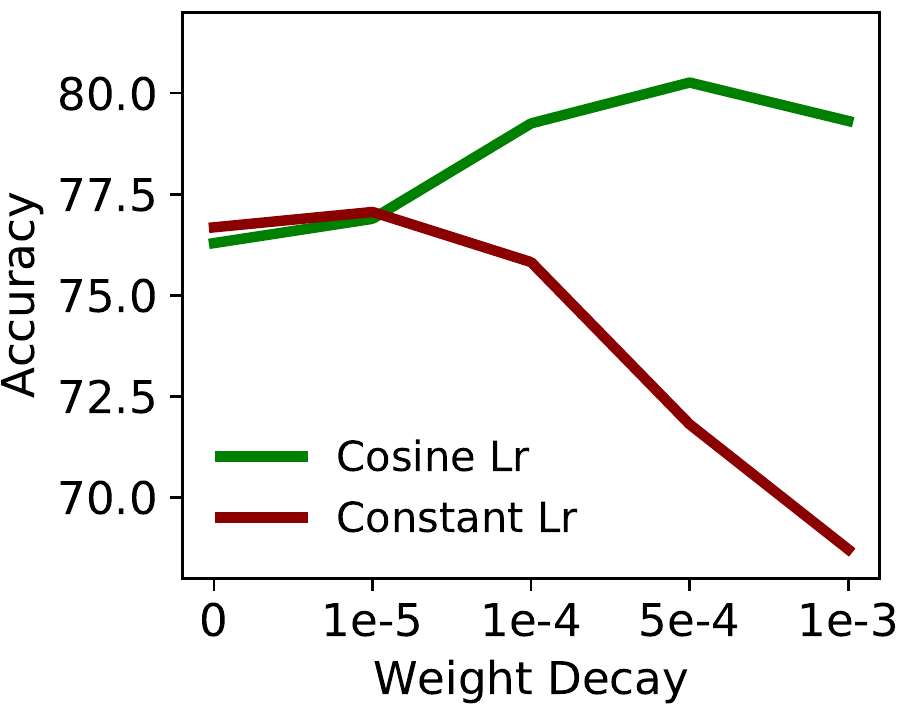} \\
    % \small (a) weights decay &  \
    % \small (b) const/cos learning rate &  \
    % \small (c) flips with constant lr. &  \
    % \small (d) flips with cosine  lr. &  \
    % \small (e) accuracy \\
        \small (a) Latent weights magnitude &  \
    \small (b) Gradient magnitude &  \
    \small (c) flips with constant lr. &  \
    \small (d) flips with cosine  lr. &  \
    \small (e) accuracy \\
    \end{tabular}
    \caption{\small \textbf{Empirical analysis (a): Effect of hyper-parameters.}
    We show the effect of weight decay and learning rate decay on binary weights flips using the Conv2 network on Cifar-10.
    Carefully tuning these hyper-parameters is important for adequately training the binary networks.
    }
     \label{fig:hyper}
\end{figure*}

%-------------------------------------------------------------------------------------------------
\begin{figure*}[t]
    \centering
    \begin{tabular}{c@{}c@{}c@{}c@{}c}
    \includegraphics[width=0.198\textwidth]{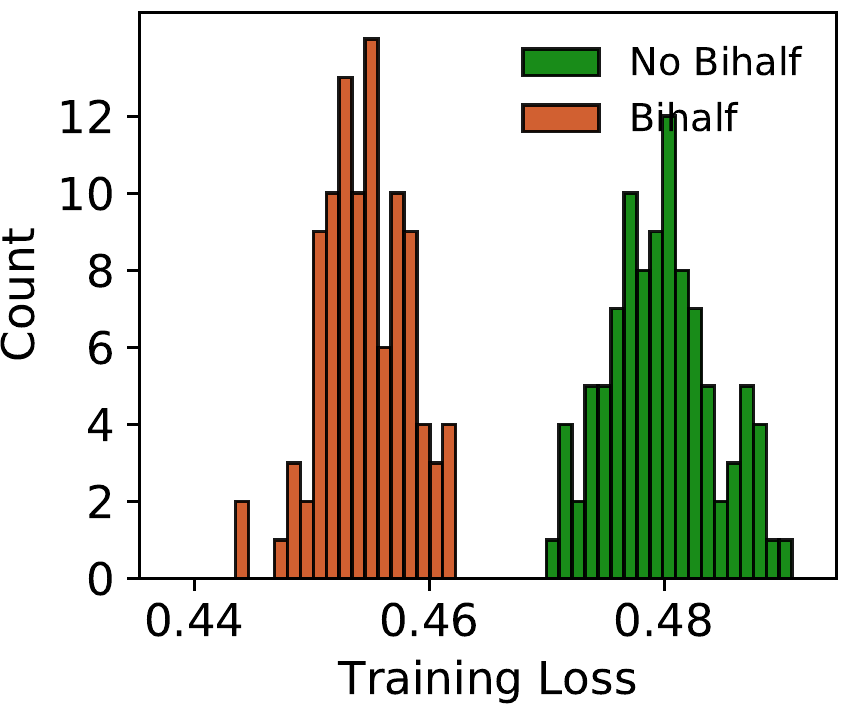} &
    \includegraphics[width=0.198\textwidth]{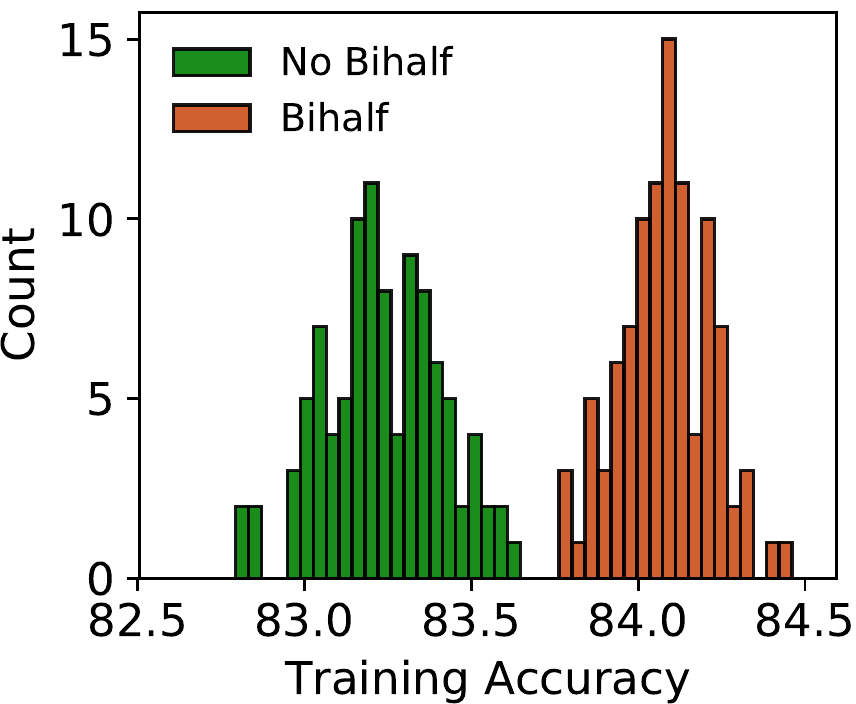} &
    \includegraphics[width=0.198\textwidth]{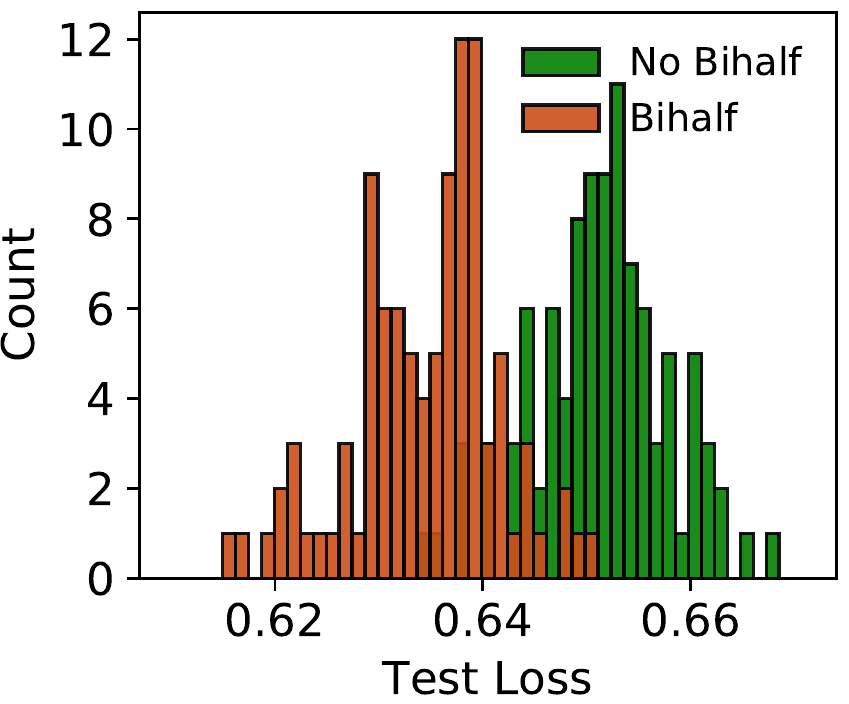} &
    \includegraphics[width=0.198\textwidth]{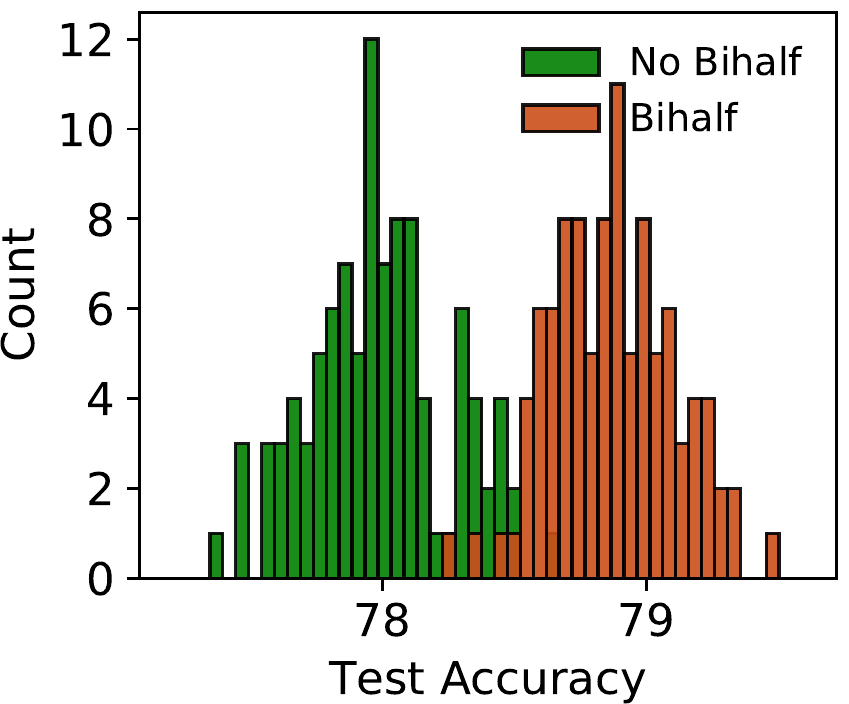} &
    \includegraphics[width=0.21\textwidth]{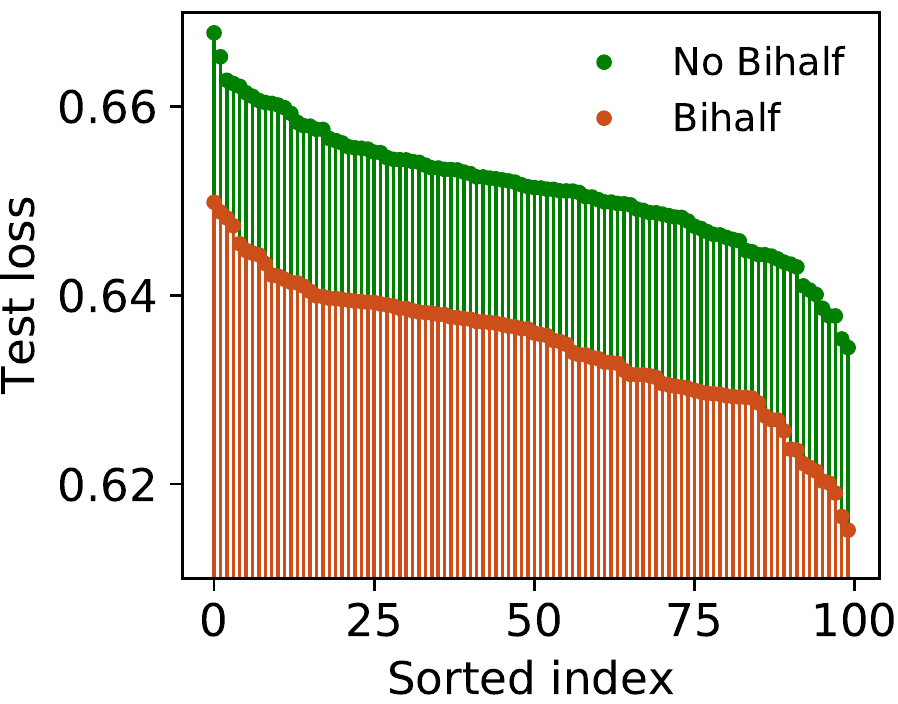} \\
    \end{tabular}
    \caption{\small \textbf{Empirical analysis (c): Optimization benefits.}
    We train our \model model 100 times on Cifar-10 and plot the distribution of the losses and accuracies over the 100 repetitions.
    We compare our results using optimal transport to the results
    using  the  standard $\sign$ function.
    % to the $\sign$ function baseline.
    On average our \model model tends to arrive at better losses and accuracies than the baseline.
    }
    \label{fig:runing100times}
\end{figure*}
\begin{figure}[t]
\centering
    \includegraphics[width=0.45\textwidth]{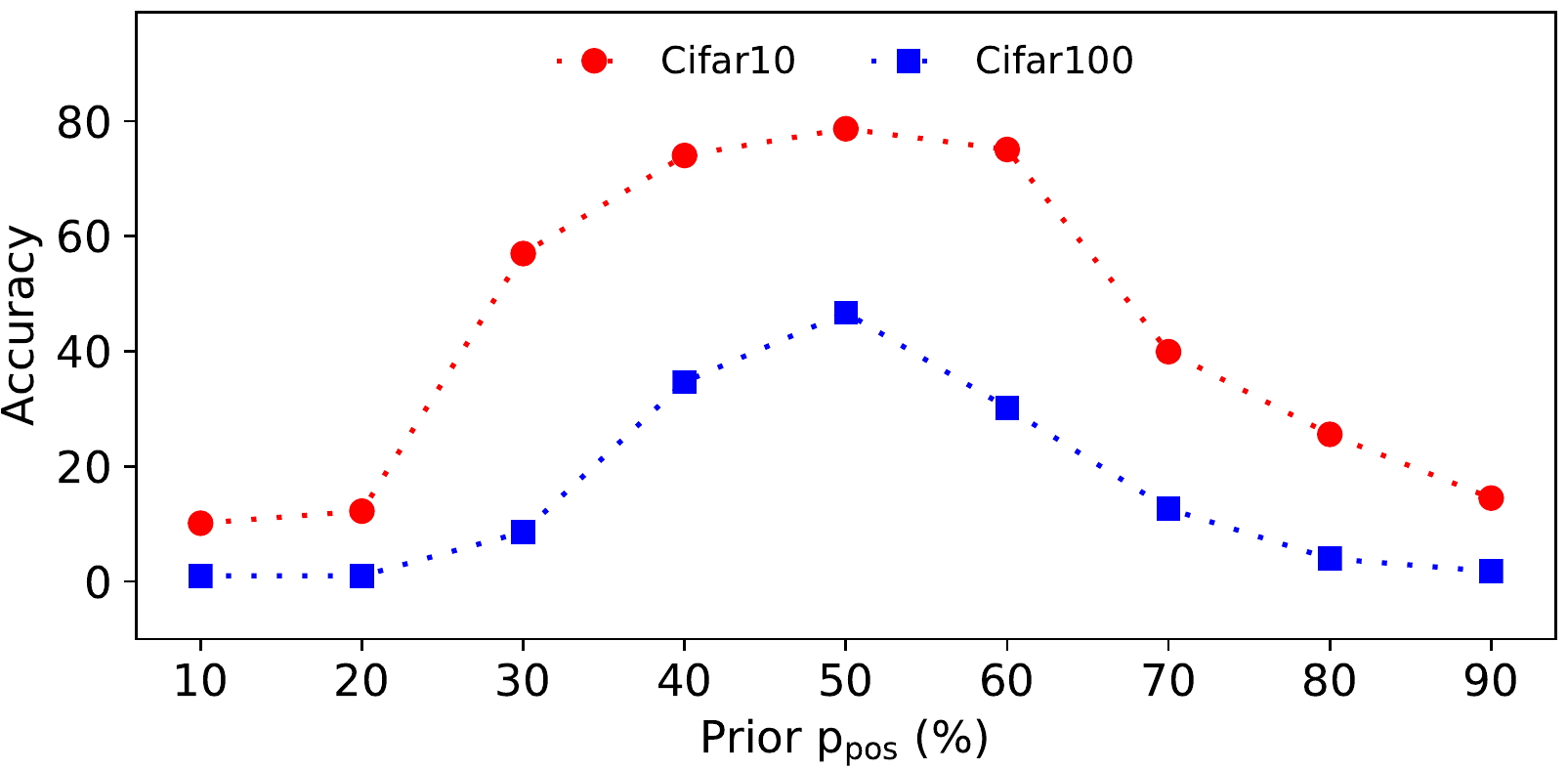}
    \caption{ \small \textbf{Empirical analysis (b): Which bit-ratios are preferred?}
    We varying the bit-ratios on Cifar-10 and Cifar-100 using Conv2 the choice of the prior$p_{pos}$ under the Bernoulli distribution. The x-axis is the probability of the $+1$ connections denoted by $p_{\text{pos}}$ in the Bernoulli prior distribution, while the y-axis denotes the top-1 accuracy values.
    Results are in agreement with the hypothesis of Qin \etal \cite{qin2020forward} that equal priors as imposed in our \model model are preferable.
    % \slp{Make it more elongated.}
    }
\label{fig:entropybitratio}
\end{figure}

\begin{figure*}[t]
\centering
\begin{tabular}{ccc}
    \includegraphics[width=0.295\textwidth]{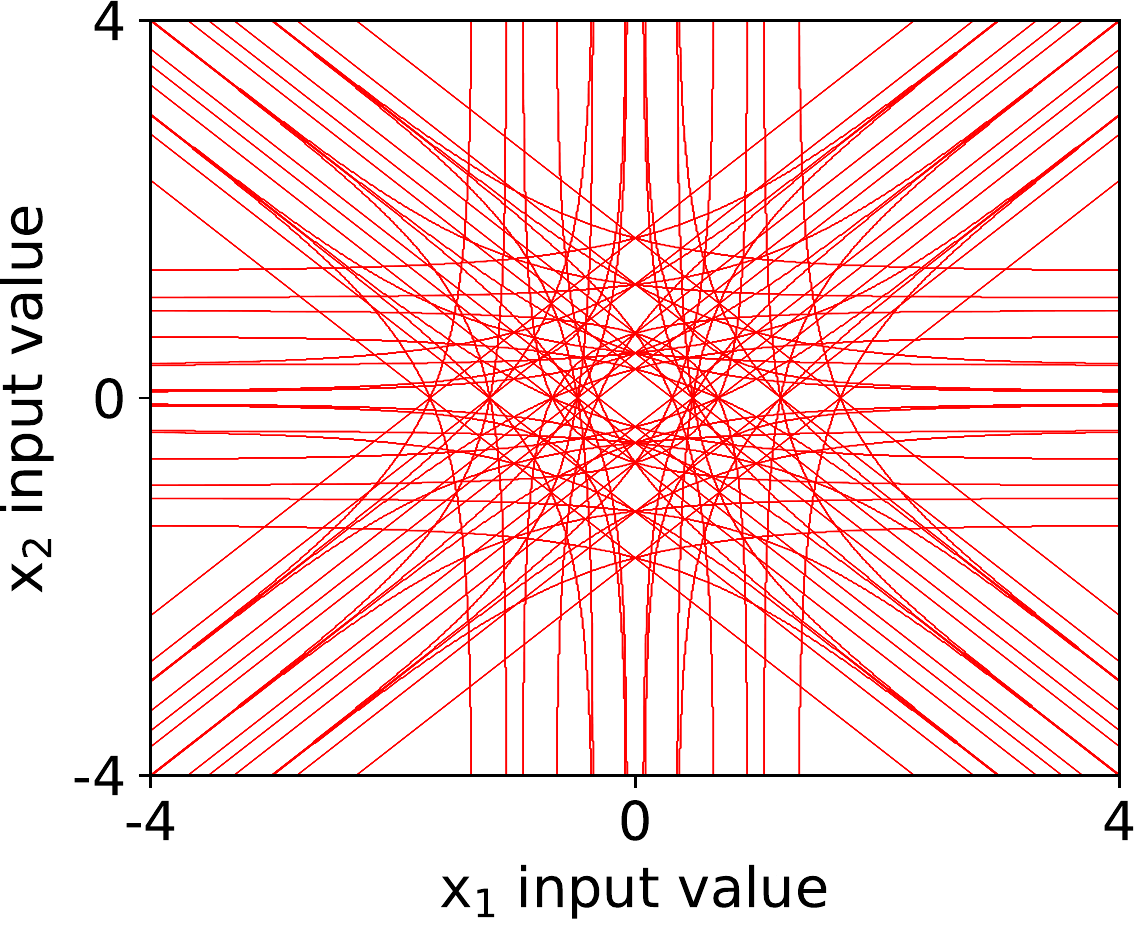} &
    \includegraphics[width=0.295\textwidth]{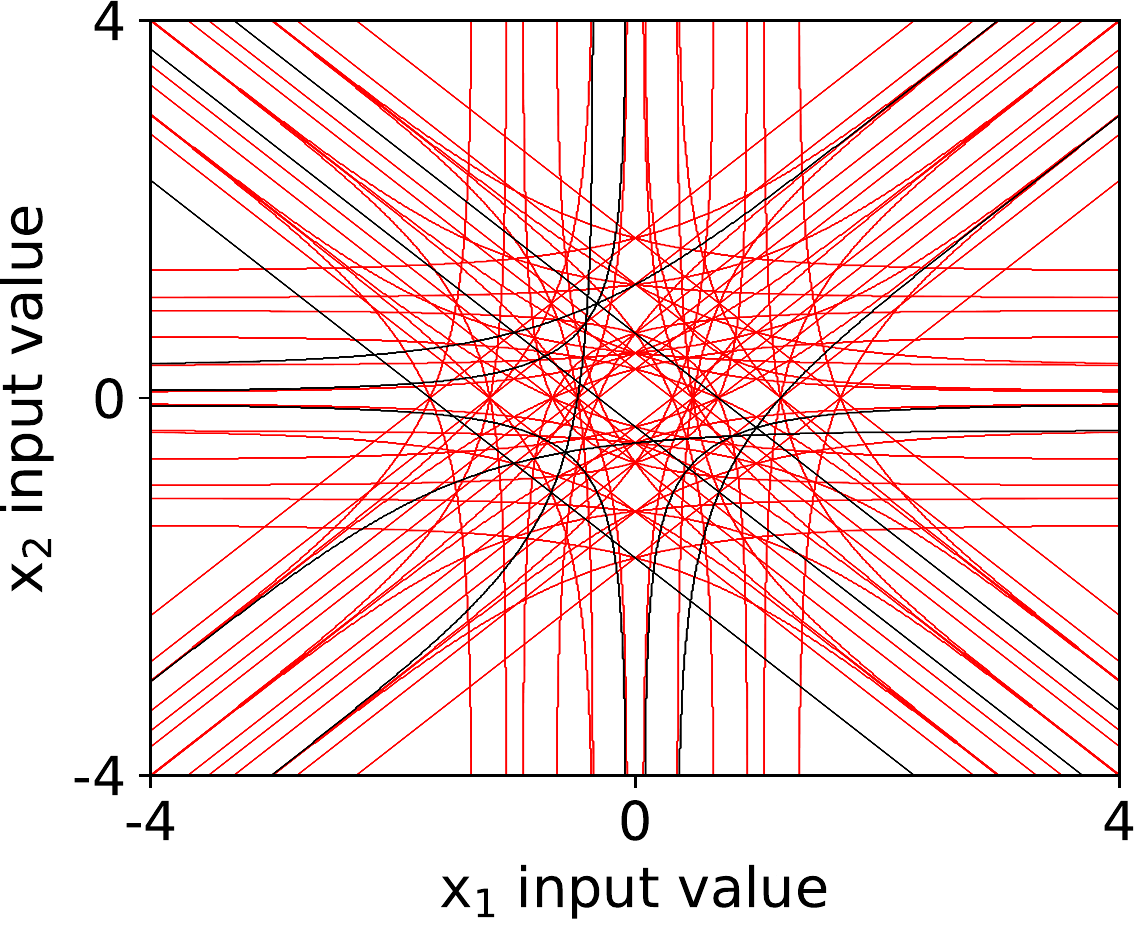} &
        \includegraphics[width=0.31\textwidth]{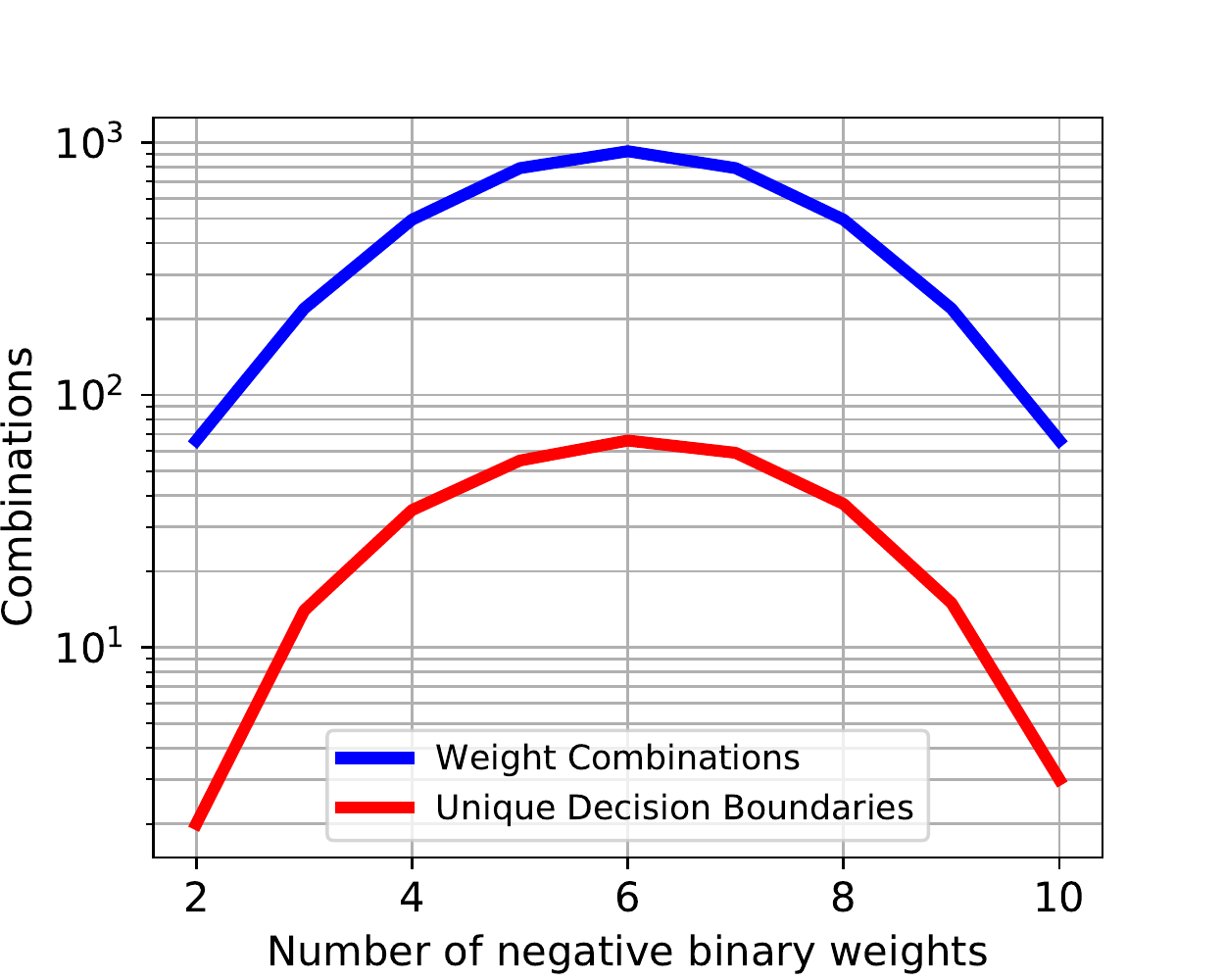}
    \\
    \scriptsize{Weight combinations: 100\% (4096).} &
    \scriptsize{Weight combinations: 22.5\% (924).} \\
    \scriptsize{Unique solutions: 100\% (76).} &
    \scriptsize{Unique solutions: 86.8\% (66).}
    &
    \scriptsize{(c) Solution space}\\
     \scriptsize{(a) Full binary network} &
    \scriptsize{(b) \Model network} &
    \scriptsize{ vs bit-ratios}
    \end{tabular}
    \caption{ \small
    \textbf{Empirical analysis (c): Optimization benefits.}
    \Model regularization:
    2D example for a 12-parameter fully connected binary network $\sigma( \mathbf{w_2}^\intercal \sigma(\mathbf{w_1}^\intercal \mathbf{x} + \mathbf{b_1}) )$,  where $\sigma(\cdot)$ is a sigmoid nonlinearity. Weights are in $\{ -1, 1 \}$.
    (a) Enumeration of all decision boundaries for 12 binary parameters (4096 = $2^{12}$ combinations).
    (b) Weight combinations and unique solutions when using our \model constraint.
    (c) The weight combinations and unique decision boundaries for various bit-ratios.
    % There are 12 binary weights in total.
    When the number of negative binary weights is 6 on the x-axis, we have equal bit-ratios, which is the optimal ratio.
    Using the bi-half works as a regularization, reducing the search-space while retaining the majority of the solutions.
    }
 \label{fig:decision_bound}
\end{figure*}

\subsection{Empirical analysis}
%-------------------------------------------------------------------------------------------

\subsubsection{(a) Effect of hyper-parameters.}
In \fig{hyper} we study the effectiveness of the commonly used training techniques of varying the weight decay and learning rate decay, when training the Conv2 network on Cifar-10.
\fig{hyper}(a) shows that using a higher weight decay reduces the magnitude of latent weights during training and therefore the magnitude of the cut-off point (threshold) between the positive and negative values.
\fig{hyper}(b) compares the gradient magnitude of two different learning rate (lr) schedules: ``constant lr" and ``cosine lr". The magnitude of the gradients reduces during training when using the cosine learning rate.
In \fig{hyper}(c) we find that increasing the weight decay for binary network with a constant learning rate schedule, increases binary weights flips.
\fig{hyper}(d) shows that decaying the learning rate when using a cosine learning rate schedule gradually decreases the number of flipped weights.
\fig{hyper}(e) shows that the choice of weight decay and learning rate decay affect each other.
Our bi-half method uses the rank of latent weights to  flip the binary weights.
A proper tuned hyper-parameter of weight decay and learning rate decay
will affect the flipping threshold.
Therefore in the experiments, we carefully tune  the  hyper-parameters of  weight decay  and  learning rate decay to build a competitive baseline.
% \jvg{For each of these 5 points: can you say why each one is important? (so what? Why are they relevant?)}
% \yq{Is this proper for the figure: "Our bi-half method uses the rank of latent weights to  flip the binary weights.
% A proper tuned hyper-parameter of weight decay and learning rate decay
% will affect the flipping threshold.
% Therefore in the experiments, we carefully tune  the  hyper-parameters of  weight decay  and  learning rate decay to build a competitive baseline. "
% % is important for updating the binary weights

% % controlling the magnitude of latent weights and gradients magnitude  can control the flipping threshold. (b).
% % The gradients magnitude determine update the flipping threshold
% }
% \jvg{sorting is independent of the magnitude, no?.. To me, these results show some insight, but its not clear to me what exactly the insights are.. but lets keep it like this.}
% \slp{Conclude what do you use in the end: We use a weight decay of 1e-4 and a learning rate schedule ...}

%-------------------------------------------------------------------------------------
\subsubsection{(b) Which bit-ratio is preferred?}
\begin{figure*}[t]
    \centering
    \includegraphics[width=0.99\textwidth]{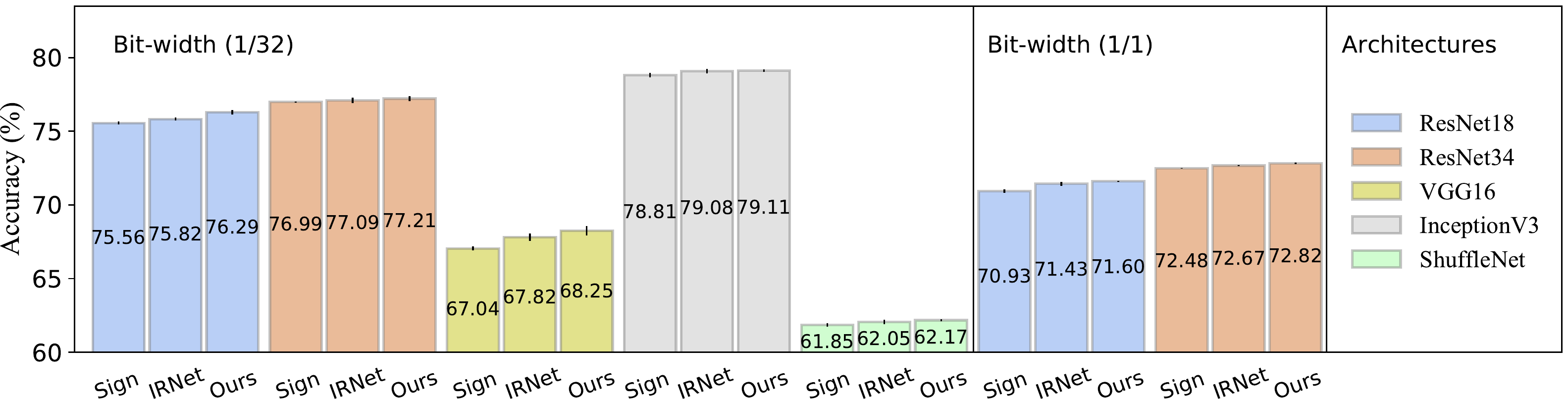}
    \caption{\small
    \textbf{Architecture variations: Different architectures on Cifar-100.}
    We evaluate on Cifar-100 over 5 different architectures: VGG16~\cite{simonyan2014very},  ResNet18~\cite{he2016deep}, ResNet34~\cite{he2016deep}, InceptionV3~\cite{szegedy2016rethinking}, ShuffleNet~\cite{zhang2018shufflenet}.
    We compare $\sign$~\cite{rastegari2016xnor}, IR-Net~\cite{qin2020forward} and our \model. The 1/32 and 1/1 indicate the bit-width for weights and for activations, where 1/1 means we quantize both the weights and the activations to binary code values.
    Our method achieves competitive accuracy across different network architectures.
    }
    \label{fig:architectures}
\end{figure*}

In \fig{entropybitratio}, we evaluate the choice of the prior $p_{\text{pos}}$ in the Bernoulli distribution for Conv2 on Cifar-10 and Cifar-100.
By varying the bit-ratio, the best performance is consistently obtained when the negative and positive symbols have equal priors as in the \model model.
Indeed, as suggested in \cite{qin2020forward}, when there is no other a-priori reason to select a different $p_{pos}$, having equal bit ratios is a good choice.

%=========================================================================================
\subsubsection{(c) Optimization benefits with bi-half.}
The uniform prior over the $-1$ and $+1$ under the Bernoulli distribution regularizes the problem space, leading to only a subset of possible weight combinations available during optimization.
We illustrate this intuitively on a 2$D$ example for a simple fully-connected neural network with one input layer, one hidden layer, and one output layer in a two-class classification setting.
We consider a 2$D$ binary input vector $\mathbf{x} = [x_1, x_2]^\intercal$, and define the network as: $\sigma( \mathbf{w_2}^\intercal \sigma(\mathbf{w_1}^\intercal \mathbf{x} + \mathbf{b_1}))$, where $\sigma(\cdot)$ is a sigmoid nonlinearity, $\mathbf{w_1}$ is a $[2 \times 3]$ binary weight matrix, $\mathbf{b_1}$ is $[3 \times 1]$ binary bias vector, and $\mathbf{w_2}$ is a $[3 \times 1]$ binary vector.
We group all $12$ parameters as a vector $\mathbf{B}$.
We enumerate all possible binary weight combinations in $\mathbf{B}$, \ie \ $2^{12}= 4096$, and plot all decision boundaries that separate the input space into two classes as shown in \fig{decision_bound}(a). All possible 4096 binary weights combinations offer only $76$ unique decision boundaries.
In \fig{decision_bound}.(b) the Bernoulli distribution over the weights with equal prior (\model) regularizes the problem space: it reduces the weight combinations to $924$, while retaining $66$ unique solutions, therefore the ratio of the solutions to the complete search spaces is increased nearly 4 times.
\fig{decision_bound}.(c)  shows in a half-log plot how the numbers of weight combinations and unique network solutions change with varying bit-ratios. Equal bit ratios is optimal.

~\\

In \fig{runing100times} we train the Conv2 networks 100 times on Cifar-10 and plot the distribution of the training and test losses and accuracies.
We plot these results when using the \model model optimization with optimal transport and by training the network using the standard $\sign$ function.
The figure shows the \model method consistently finds better solutions with lower training and test losses and higher training and test accuracy.
To better visualize this trend we sort the values of the losses for our \model and the baseline $\sign$ method over the 100 repetitions and plots them next to each other.
On average the \model finds better optima.

\subsection{Architecture variations}

In \tab{convnetacc} we compare the $\Sign$~\cite{rastegari2016xnor}, IR-Net~\cite{qin2020forward} and our \model on four shallow Conv2/4/6/8 networks on Cifar-10 (averaged over 5 trials).
As the networks become deeper, the proposed \model method consistently outperforms the other methods.

In \fig{architectures}, we further evaluate our method on Cifar-100 over 5 different architectures: VGG16~\cite{simonyan2014very},  ResNet18~\cite{he2016deep}, ResNet34~\cite{he2016deep}, InceptionV3~\cite{szegedy2016rethinking}, ShuffleNet~\cite{zhang2018shufflenet}. Our method is slightly more accurate than the other methods, especially on the VGG16 architecture, it never performs worse.

\begin{figure*}[t]
    \centering
    \begin{tabular}{c@{}c@{}c@{}c@{}c}
    \includegraphics[width=0.19\textwidth]{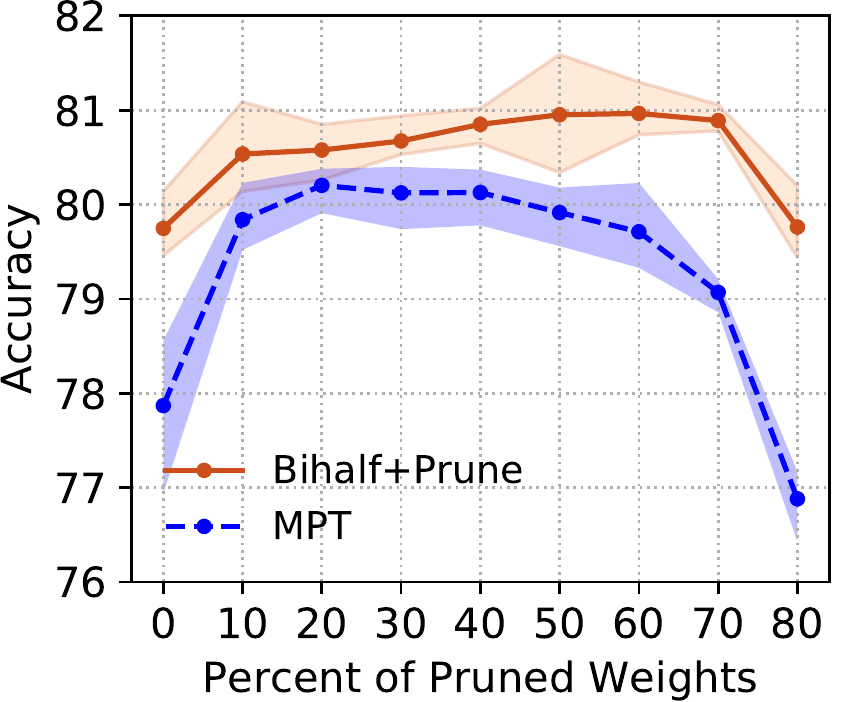} &  \  \
    \includegraphics[width=0.19\textwidth]{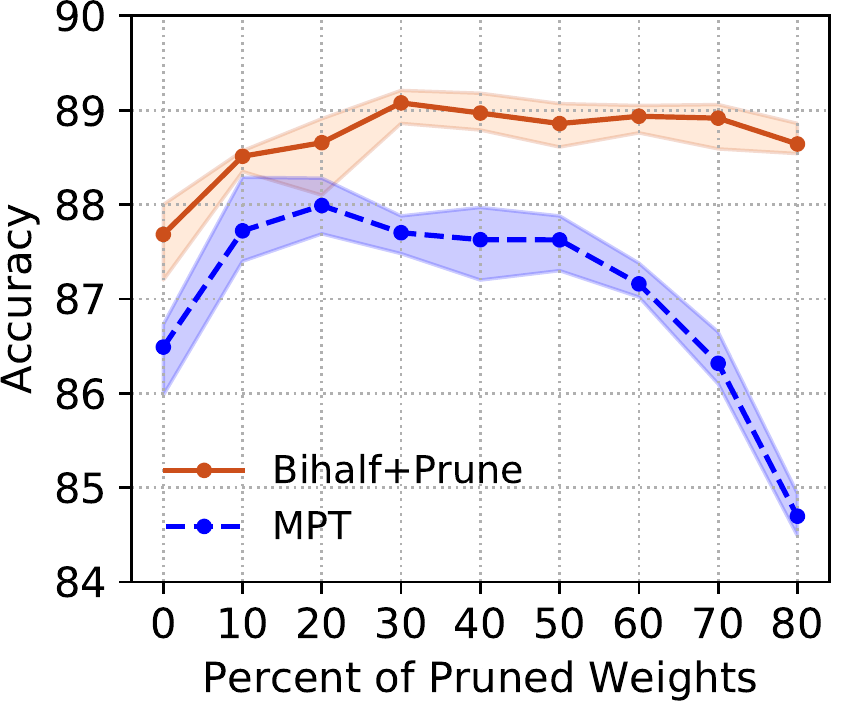} &    \  \
    \includegraphics[width=0.19\textwidth]{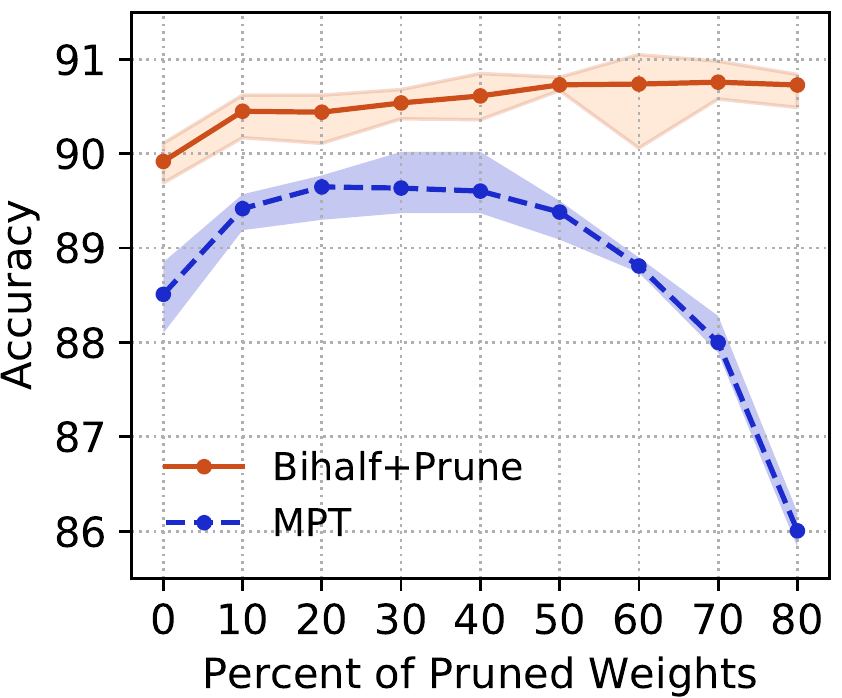} &    \  \
    \includegraphics[width=0.19\textwidth]{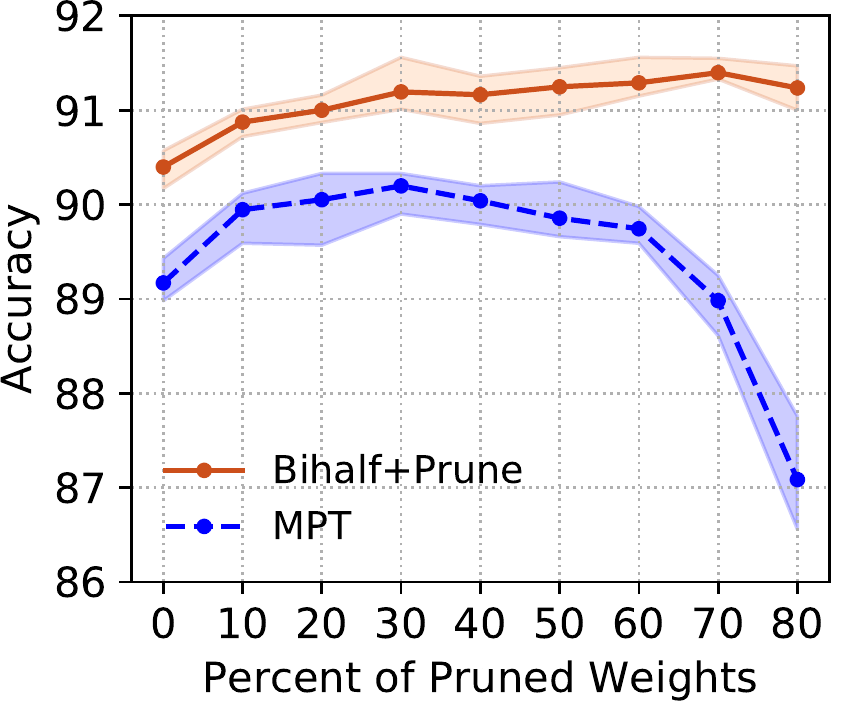} &   \  \
    \includegraphics[width=0.199\textwidth]{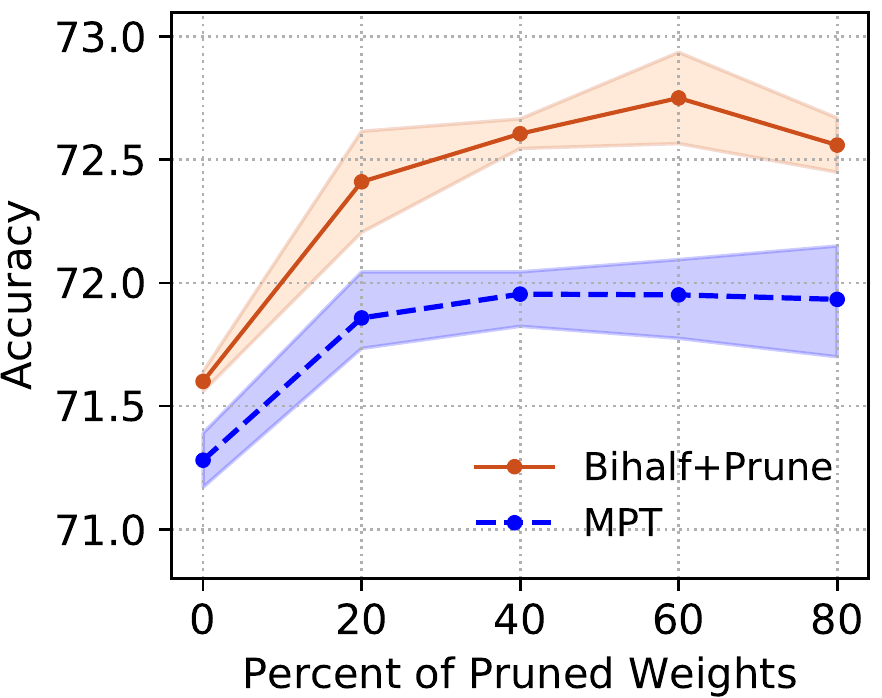} \\
        \ \ \ \ \ \small{Conv2} & \ \  \ \ \ \ \ \small{Conv4} & \ \ \ \ \ \ \ {Conv6}& \ \ \ \ \ \ \ \small{Conv8}  & \ \ \ \ \ \ \ \small{ResNet-18}  \\
    \end{tabular}
    \caption{\small
    \textbf{Comparison with state-of-the-art (b): Pruned networks.}
    Test accuracy of Conv2/4/6/8 on CIFAR-10, and ResNet-18 on CIFAR-100 when varying the \% pruned weights.
    We compare with the MPT baseline \cite{diffenderfer2021multi} using binary weight masking and the $\sign$ function.
    Having equal $+1$ and $-1$ ratios is also optimal when the networks rely on pruning and that our optimal transport optimization can easily be adapted to work in combination with pruning.
    }
    \label{fig:prune}
\end{figure*}

\begin{table}[t]
	\centering
	\renewcommand{\arraystretch}{1.2}
	\resizebox{1\linewidth}{!}{
	 \setlength{\tabcolsep}{0.95mm}{
	   	\begin{tabular}{lcccccc}
    	\toprule
		 Method & Conv2  & Conv4 &  Conv6  & Conv8 \\\midrule		
		 $\Sign$ & 77.86$\pm$ 0.69  & 86.49$\pm$ 0.24 &  88.51$\pm$ 0.35 &  89.17$\pm$0.26  \\
		 IR-Net & 78.32$\pm$ 0.25  & 87.20$\pm$ 0.26 &  89.61$\pm$ 0.11 &  90.06$\pm$0.06  \\ 		
		 \Model (ours) & 79.25$\pm$ 0.28  & 87.68$\pm$ 0.32 &  89.92$\pm$ 0.19 &  90.40$\pm$0.17  \\	
	   \bottomrule
	\end{tabular}
	}}
	\caption{\small \textbf{Architecture variations.}
	Accuracy comparison of $\sign$~\cite{rastegari2016xnor}, IR-Net~\cite{qin2020forward} and our \model on Conv2/4/6/8 networks using Cifar-10, over 5 repetitions.
	As the depth of the network increases, the accuracy of our method increases.
    }
	\label{tab:convnetacc}
\end{table}

%--------------------------------------------------------------------------------------------
\begin{table}[t]
	\centering
	    \renewcommand{\arraystretch}{1.2}
		\setlength{\tabcolsep}{1mm}{
	    \resizebox{0.95\linewidth}{!}{
		\begin{tabular}[c]{llcccclccccc}
		\toprule			
        \multirow{1}*{Backbone}  &  \multirow{1}*{Method}&    \multirow{1}*{Bit-width} &   Top-1(\%) & {Top-5(\%) } \\
        & &  (W/A) \\\midrule
		\multirow{10}*{ResNet-18}  & FP & 32/32  &  69.3 &  89.2  \\ % \cmidrule(l){2-5}
    	& ABC-Net & 1/1 &   42.7 &  67.6      \\	
		& XNOR & 1/1  &    51.2&  73.2    \\ 	
		& BNN+& 1/1 &  53.0 & 72.6   \\
		& Least-squares  & 1/1 & 58.9 & 81.4 \\
				& XNOR++ & 1/1 &  57.1 & 79.9  &  \\
		& IR-Net & 1/1 &  58.1 & 80.0  \\
		& RBNN  & 1/1 &   59.9 &   {81.9 }  \\	
% 		& Real2Binary & 1/1 &  60.9 & 83.0  \\
        & $\Sign$ (Baseline) & 1/1 &  59.98 & 82.47  \\
		& \Model (ours) & 1/1 &  \textbf{60.40}  &   \textbf{82.86}           \\	
	    \midrule
		\multirow{6}*{ResNet-34}  & FP & 32/32  &   73.3 & 91.3   \\ %\cmidrule(l){2-5}
	    &	ABC-Net & 1/1  & 52.4 & 76.5 \\
        & Bi-Real&  1/1 & 62.2 & 83.9 \\
		& IR-Net  & 1/1 &   {62.9} &   {84.1}  \\
	    & RBNN  & 1/1 &   63.1 &   {84.4 }  \\	
        & \model (ours)  & 1/1 & \textbf{64.17} &  \textbf{85.36}    \\
        \bottomrule
		\end{tabular}}}
		\vspace{0.05in}
	\caption{\small
	\textbf{Comparison with state-of-the-art (a): ImageNet results.}
	We show Top-1 and Top-5 accuracy on ImageNet for a number of state-of-the-art binary networks. $\Sign$ is our baseline by  carefully tuning the hyper-parameters.
	Our proposes \model model consistently outperforms the other binarization methods on this large-scale classification task.
% 	\slp{What is $\Sign$ (ours)? We cannot do sign cause we do optimal transport. This is "ours"}
    }
\label{tab:imagnet}
\end{table}

%====================================================================================
\subsection{Comparison with state-of-the-art}
%-----------------------------------------------------------------------------------------
\subsubsection{(a) Comparison on ImageNet.}
For the large-scale ImageNet dataset we evaluate a ResNet-18 and ResNet-34 backbone \cite{he2016deep}.
\tab{imagnet} shows a number of state-of-the-art quantization methods over ResNet-18 and ResNet-34, including:
ABC-Net~\cite{lin2017towards}, XNOR~\cite{rastegari2016xnor},
BNN+~\cite{darabi2018bnn+},
Bi-Real~\cite{liu2018bi},
RBNN~\cite{lin2020rotated},
XNOR++~\cite{bulat2019xnornet},
IR-Net~\cite{qin2020forward}, and
Real2binary \cite{martinez2020training}.
Of all the methods, RBNN is the closest in accuracy to our \model model.
This is because RBNN relies on the $\sign$ function but draws inspiration from hashing, and adds an activation-aware loss to change the distribution of the activations before binarization.
On the other hand, our method uses the standard classification loss but outperforms most other methods by a large margin on both ResNet-18 and ResNet-34.

%---------------------------------------------------------------------------------------------
\subsubsection{(b) Comparison on pruned networks.}

In \fig{prune} we show the effect of our \model on pruned models.
Following the MPT method \cite{diffenderfer2021multi} we learn a mask for the binary weights to prune them.
However, in our \model approach for pruning we optimize using optimal transport for equal bit ratios in the remaining unpruned weights.
We train shallow Conv2/4/6/8 networks on CIFAR-10, and ResNet-18 on CIFAR-100 while varying the percentage of pruned weights.
Each curve is the average over five trials.
Pruning consistently finds subnetworks that outperform the full binary network.
Our \model method with optimal transport retains the information entropy for the pruned subnetworks, and consistently outperforms the MPT baseline using the $\sign$ function for binarization.

\section{Conclusion}
We focus on binary networks for their well-recognized efficiency and memory benefits.
To that end, we propose a novel method that optimizes the weight binarization by aligning a real-valued proxy weight distributions with an idealized  distribution using optimal transport. 
This optimization allows to test which prior bit ratio is preferred in a binary network, and we show that the equal bit ratios, as advertised by \cite{qin2020forward}, indeed work better.
We confirm that our optimal transport binarization has optimization benefits such as: reducing the search space and leading to better local optima. 
Finally, we demonstrate competitive performance when compared to state-of-the-art, and improved accuracy on 3 different datasets and various architectures.
We additionally show accuracy gains with pruning techniques. 

% \input{APPENDIX}
% {\small
% \bibliographystyle{ieee_fullname}
% \bibliography{egbib}
% }
%\clearpage
	{
		\bibliographystyle{aaai}
		\bibliography{main}
	}
%\appendix
%\section{Appendix}
%You may include other additional sections here.
% \clearpage
% \clearpage
% \appendix
%  \input{APPENDIX}
\end{document}